\documentclass[lettersize,journal]{IEEEtran}
\usepackage{amsmath,amsfonts}
\usepackage{algorithmic}
\usepackage{algorithm}
\usepackage{array}
\usepackage[caption=false,font=normalsize,labelfont=sf,textfont=sf]{subfig}
\usepackage{textcomp}
\usepackage{stfloats}
\usepackage{url}
\usepackage{verbatim}
\usepackage{graphicx}
\usepackage{cite}
\usepackage{verbatim}
\usepackage{amsmath}
\usepackage{amssymb}
\def\Set#1{\mathbb#1} 
\def\V#1{\boldsymbol{\mathit#1}}          
\def\Cal#1{\mathcal#1}

\newtheorem{theorem}{Theorem}

\newtheorem{definition}{Definition}

\usepackage{multirow}

\begin{document}

\title{Dual Path Structural Contrastive Embeddings for Learning Novel Objects}

\author{
	
Bingbin Li, Elvis Han Cui, Yanan Li, \IEEEmembership{Member,~IEEE}, Donghui Wang, \IEEEmembership{Member,~IEEE}, Weng Kee Wong, \IEEEmembership{Member,~IEEE}
\thanks{This paper was produced by the IEEE Publication Technology Group. They are in Piscataway, NJ.}
\thanks{Manuscript received April 19, 2021; revised August 16, 2021. }
\thanks{Bingbin Li is with the Department of Computer Science and Technology, Zhejiang University, Hangzhou 310027, China (e-mail: bingbin\_lee@zju.edu.cn).}
\thanks{Elvis Han Cui is with the Department of Biostatistics, University of California, Los Angeles, Los Angeles, 90024, CA, USA (e-mail: elviscuihan@g.ucla.edu). }
\thanks{Yanan Li is with the Research Center for Applied Mathematics and Machine Intelligence, Research Institute of Basic Theories,  Zhejiang Laboratory, Hangzhou 311100, China (e-mail: liyn@zhejianglab.com).}
\thanks{Donghui Wang is with the Department of Computer Science and Technology, Zhejiang University, Hangzhou 310027, China (e-mail: dhwang@zju.edu.cn).}
\thanks{Weng Kee Wong is with the Department of Biostatsitics, University of California, Los Angeles, Los Angeles, CA 90095-1772, USA (e-mail: wkwong@ucla.edu). }
}

\markboth{Journal of \LaTeX\ Class Files,~Vol.~14, No.~8, August~2021}%
{Shell \MakeLowercase{\textit{et al.}}: A Sample Article Using IEEEtran.cls for IEEE Journals}


\maketitle

\begin{abstract}
	
Learning novel classes from a very few labeled samples has attracted increasing attention in machine learning areas.  Recent research on either meta-learning based or transfer-learning based paradigm demonstrates that gaining information on a good feature space can be an effective solution to achieve favorable  performance on few-shot tasks. In this paper, we propose a simple but effective paradigm that decouples the tasks of learning feature representations and classifiers and only learns the feature embedding architecture from base classes via the typical transfer-learning training strategy. To maintain both the generalization ability across base and novel classes and discrimination ability within each class, we propose a dual path feature learning scheme that effectively combines structural similarity with contrastive feature construction. In this way, both inner-class alignment and inter-class uniformity can be well balanced, and result in improved performance. Experiments on three popular benchmarks show that when incorporated with a simple prototype based classifier, our method can still achieve promising results for both standard and generalized few-shot problems in either an inductive or transductive inference setting.

\end{abstract}


\section{Introduction}
\IEEEPARstart{C}{onvolutional} neural network (CNN) based models have achieved significant advances in various computer vision tasks,  such as object recognition \cite{russakovsky2015imagenet}, object detection \cite{ren2015faster}, semantic segmentation \cite{he2017mask}, etc. These successes can be attributed to both the advancement of deep models and the availability of large-scale labeled datasets. However, in many real-world scenarios, there are only a limited number of labeled samples, e.g. medical imaging and manufacturing, due to the prohibitively costly acquisition.  In this case, the straightforward application of deep learning models would suffer from severe overfitting and model bias, thus leading to considerate performance degradation. In contrast, humans can rapidly learn novel concepts after observing only one or a few instances. Such ability to learn from few examples is desirable for above modern models in a low-data regime. To narrow down this gap between machine and human learning, few-shot learning (FSL) is emerging as an appealing paradigm and receiving increasing research interests \cite{NIPS2016_90e13578,snell2017prototypical,finn2017modelagnostic,chen2020closer}. 

\begin{figure}[!htbp]
	\centering
	\includegraphics[width=0.5\textwidth]{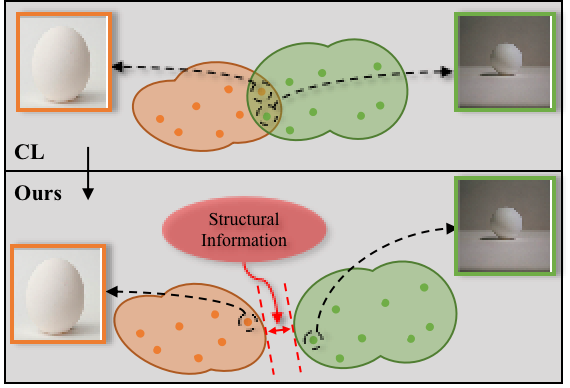}
	\caption{Some categories may have similar visual features and dissimilar semantics. Naively applying contrastive learning (CL) in FSL, the learned features of these categories would be gathered around. We incorporate structural similarity within the contrastive feature construction to improve both discrimination and generalization of these features. }
	\label{fig: motivation-diagram}
\end{figure}

In contrast to conventional supervised learning, there are large amounts of labeled data for base classes and only a few labeled samples for disjoint novel classes available in FSL. In order to learn a well-performed classifier of novel classes, one general strategy is to first learn some transferable prior knowledge on base classes and then adapt the knowledge quickly to the novel ones. In terms of the generalization strategy, current FSL models can be roughly divided into three groups. (1) Optimization-based models \cite{li2017meta, finn2017modelagnostic,nichol2018first} generally learn good initial model parameters from base classes, so that they can be quickly updated via one or a few gradient steps to achieve optimum that can provide a good performance on the novel task.  Representatives are Model-Agnostic Meta-Learning (MAML) \cite{finn2017modelagnostic} and its variants \cite{jeong2020ood}.  (2) Metric-based models \cite{snell2017prototypical,tian2020rethinking}  aim at learning a good embedding model that can transform input samples into an appropriate metric spaces, where a simple linear classifier based on certain specific measures can perform well on novel task. On top of the learned embeddings, \cite{simon2020adaptive, yoon2019tapnet} further explore adaptive classifiers that are specific to each novel task for performance improvement. (3) Hallucination-based models learn to augment support samples for each novel classes via generative models \cite{zhang2019few, li2020adversarial}, pseudo-labeling \cite{li2019learning, wang2020instance}, so that the low-shot problem could be alleviated by some extent.

Among all these above methods, a large body of works focus on tackling the FSL problem using the idea of meta-learning via episodic-training strategy. It randomly constructs a set of fake few-shot tasks that mimic the real testing scenario from base classes \cite{hospedales2020meta}. Each few-shot task (usually a N-way K-shot task) consists of limited support samples, 
which are used for meta-learner construction, and a bunch of query samples for evaluation. Through training on a series of fake tasks, the model is expected to grasp the capability of learning to learn, so that it could be quickly adapted to the real novel classes via several labeled samples. Recent works have demonstrated that powerful feature representations learned by episodic-training strategy from bases classes can outperform current sophisticated FSL algorithms. \cite{raghu2019rapid} shows that the meta-initialization  in MAML already provides high quality representations and can be used without adaptation to perform unseen tasks. \cite{chen2020closer, dhillon2019baseline} demonstrates that a baseline feature extractor trained on all meta-train set can already achieve comparable performance. At the same time, another parallel transfer learning strategy is gaining increasing attention. Different from the meta-learning paradigm, it directly fine-tunes the pre-trained model from base classes on the novel task. Empirical studies show that this simple transfer learning strategy can achieve on-par or even better performance that meta-learning algorithms \cite{huang2019all, tian2020rethinking, wang2020frustratingly}. For example, \cite{huang2019all} finds that fine-tuning only the last layer of existing detectors on rare classes is crucial to the few-shot object detection task. It outperforms the meta-learning competitors by roughly 2-20 points on current benchmarks. \cite{tian2020rethinking} discusses that whether it is the episodic-training strategy or the learned embedding space that is responsible for the success of FSL. They finds that using a good embedding model can be more effective than current meta-learning algorithms. These works demonstrated that learning a good embedding helps to improve the performance of few-shot learning under both meta-learning and transfer-learning framework.

Based on the observation, other recent works try to explore intricate strategies from unsupervised learning and semi-supervised learning within the above framework, in order to learn a better embedding model that can be quickly generalized.  For example, \cite{gidaris2019boosting} adds two kinds of self-supervision, i.e. rotation prediction and relative patch location prediction, as auxiliary tasks in the few-shot learning pipeline and shows that the feature extractor can learn richer and more transferable visual representations, thus boosting FSL performance. \cite{chen2021pareto} further takes into consideration the conflicting objectives between different auxiliary tasks and proposes to find a Pareto solution by explicitly casting the learning process into multi-objective optimization problem. \cite{li2019learning} embeds the self-training idea of semi-supervised learning into the meta gradient descent paradigm to learn fast adapted representations. \cite{lee2021unsupervised} develops a novel unsupervised adaptation scheme with feature reconstruction and dimensionality-driven early stopping that can find generalizable features. However, in learning feature embedding models, these methods either use the sophisticated meta-learning strategy or take advantage of extra auxiliary unlabeled data. This brings in an interesting question: can the feature model in FSL be pushed by simply using the traditional transfer-learning strategy without introducing any auxiliary data?

To answer this question, we propose a novel feature learning strategy  by taking inspiration from the recent contrastive learning research. It does not use any information from either auxiliary data or novel classes during training. And still it is capable of  learning a generalized feature embedding space, without any model modification for each novel class. First, we take a look at the inductive bias when directly applying the basic idea of contrastive learning in FSL. Generally speaking, contrastive learning assumes a way to sample positive pairs, representing similar samples that should have similar representations. Thus, even without any supervision,  generalized instance-aware feature vectors could be learned, when two homologous images are utilized. However, when naively applying the above idea in FSL problem, visually similar but semantically different images would gather around, due to the absence of class labels during training. For example, the ``Ping-Pong Ball'' and ``egg''  in Fig.\ref{fig: motivation-diagram} would lie close to each other in the feature embedding space. Thus,  the discrimination ability of the learned features would be degraded. To insert class semantics into the feature learning process, we could naturally take all images from the same class of an anchor as positives and the rest as negatives. However, the learned class-aware features can only provide information for classifying each base category.  As a result, even if the feature space can provide a well-separated decision boundary, it may lose the generalization  ability when novel classes are involved, since inter-class relationship is ignored in the process.

\begin{figure*}[t]
	\centering
	\includegraphics[width=\textwidth]{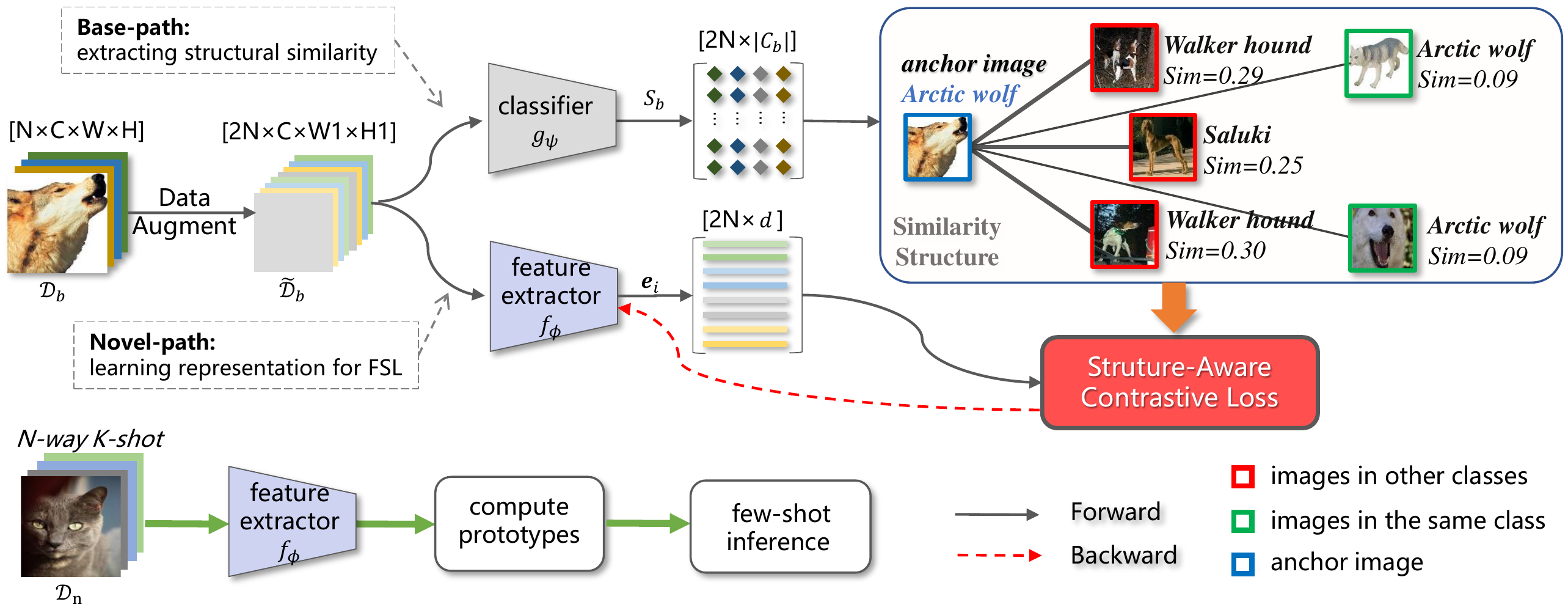}
	\caption{Illustration of our proposed dual-path FSL framework. It includes a) \textbf{base-path}, which uses  a standard CNN classifier $g_\psi$ to provide structural similarity information among all base classes, and b) \textbf{novel-path}, which incorporates structural similarities within contrastive feature construction to learn a more generalized $f_{\phi}$, without any information from novel classes.  }
	\label{fig: dualpipe}
\end{figure*}

In this paper, we propose a dual path feature model that enables the feature space in FSL to be both generalized across base and novel classes and discriminative within each class, as is shown in Fig.\ref{fig: dualpipe}.  The key is to combine contrastive feature construction with global structural information, provided by a teacher network in the base path. Each image will adaptively be contextualized over all images including both homologous and non-homologous from all base classes. In this way, the learned feature space can increase both inner-class alignment and inter-class uniformity, thus improving FSL performance. To be specific, we first train a standard deep classifier on base classes in the teacher path, which provides structural similarity information among images and is kept fixed in the next. Then, we optimize the feature learning path by the proposed structure-aware contrastive loss. For each anchor image, it treats all other images in a batch as positive samples, where each pair is given a fake similarity by the teacher path.  After training, we keep the feature learning path fixed and directly use it as the feature extractor for novel classes. We demonstrate that the trained extractor plus a simple class prototype based classifier can beat state-of-the-art performance.

In summary, our contributions are three-folds: 
\begin{itemize}
	\item We propose a simple dual path FSL method that decouples the learning of feature embeddings and classifiers. In contrast to others, it is trained on only base samples via direct transfer learning strategy, without any information from novel ones during training.  
	
	\item We propose a novel structure-aware contrastive embedding strategy. It incorporates structural similarity information among all image pairs into the contrastive feature construction process, in order to learn a generalized feature space.  We theoretically prove that it can increase inner-class alignment and inter-class uniformity. 
	
	\item We conduct extensive experiments on three benchmarks and show that when combined with a simple prototype based classifier, our method can achieve promising results for both standard and generalized FSL \cite{li2019few,ye2020fewshot}, in either inductive or transductive setting. 	
	
\end{itemize}

\section{Related Work}
\subsection{Few-Shot Learning}
Due to the high cost of collecting and annotating large-scale training data, few-shot learning (FSL) has attracted increasing attention in recent years and made considerable progress up to now. Current FSL methods can be roughly categorized into three groups: optimization-based models, metric-based models and hallucination-based models. 


%
\subsubsection{Optimization-based models}
Another line of research work follows the idea of obtaining a general optimization process over multiple episodes  within the meta-learning framework. MAML is a representative method, which finds an initialization of the neural networks that can be quickly adapted to the novel task withing one or a few update steps \cite{finn2017modelagnostic}. \cite{grant2018recasting} further reformulates MAML within a Bayesian framework to conduct probabilistic inference.  To further simplify MAML, \cite{nichol2018first} proposes a first-order gradient-based meta-learning approach Reptile, which points out that MAML can be simply implemented by removing re-initialization for each task. \cite{rusu2018meta} proposes that it is beneficial to decouple the optimization-based meta-learning algorithms from high-dimensional model parameters. It uses an encoder-decoder architecture, where the encoder is for exploring the low-dimensional latent embedding space and the decoder for predicting high-dimensional parameters. While  \cite{lee2019meta} replaces the linear predictor with an SVM in the MAML framework, by it incorporating a differentiable quadratic programming (QP) solver to allow end-to-end learning.

\subsubsection{Metric-based models}
Metric-based models have become a popular research direction in few-shot learning. With the help of novel regularizers or some extra transforming networks, embeddings learned from base classes can generalize well on novel classes \cite{hu2021leveraging, 9413783, 9412076, Yang_2020_CVPR, pmlr-v119-ziko20a, ye2020fewshot}. For example, \cite{Yang_2020_CVPR} and \cite{9412076} use graph networks to switching the information among features so as to improve the quality of embeddings.  \cite{pmlr-v119-ziko20a} utilizes Laplacian regularization to aggregate the features with high similarity. \cite{ye2020fewshot} introduces transformer in FSL to enhance the representing ability of extracted features. On top of these learned features, some methods further learn adaptive classifiers on novel classes \cite{Sung_2018_CVPR, snell2017prototypical, NEURIPS2020_1cc8a8ea, Lifchitz_2019_CVPR}. For instance, \cite{Lifchitz_2019_CVPR} explores the role of local information in FSL and trains a dense classifier. \cite{NEURIPS2020_1cc8a8ea} proposes that intervention towards the structural causal model of few-shot learning can help improve the generalization ability of networks. While \cite{li2020boosting} introduces task-relevant adaptive margin loss to increase the border distance between two similar classes. 

\subsubsection{Hallucination-based models}
The key challenge of few-shot learning lies in insufficient training data. Thus, a straightforward solution to tackle this situation is increasing the number of training data. We can apply simple data augmentation on limited training set, such as flipping, cropping and color jittering operations. However, these skills cannot significantly improve FSL performance, since no extra semantic information is added. Recent research works try to use deep neural networks to generate high quality samples based on a few training images in input space \cite{8578858, 2020DAWSON, 2020Matchinggan, 2020Few}. For example, \cite{2020Few} tunes part of the network parameters to keep the old knowledge while adapting to the new samples.  \cite{2020Matchinggan} fuses some images belonging to the same category for new image generation. 
Considering that samples generated in input space are usually unstable, another type of works  try to generate new samples in the latent feature space \cite{8237590, 2018Delta, 2017Data, 2018Low}. For example, \cite{2018Delta} uses auto-encoders to encode the abundant semantic information contained in seen categories and then transfers these information to generate novel ones. Following the similar idea, \cite{2018Delta} differently treats the covariance matrices as the transferable information in feature generation. Although these above methods can improve the performance of few-shot methods, synthesized data might contain artifacts and miss the necessary details of real data, which is the limitation of this kind of methods. Besides, how to ensure the diversity of generating samples under few-shot circumstance is still a big challenge.

A large portion of these works are based on meta-learning, in which training is done on a series of fake few-shot classification tasks (i.e. episodes) that mimics the real testing scenario. Several recent studies have questioned the necessity of meta-learning mechanism and have shown that fine-tuning a pre-trained embedding network trained on the base dataset with standard cross-entropy loss can still achieve promising results \cite{huang2019all, tian2020rethinking, wang2020frustratingly, ziko2020laplacian}. Our method in this paper lies in this line of research that doesn't require meta-learning nor novel samples during training a feature embedding model.

\subsection{Contrastive Learning }
Contrastive learning (CL) has recently achieved great success in self-supervised tasks, where homologous samples are first produced and then contrastive objective is applied to perform instance discrimination as a pretext task \cite{liu2020selfsupervised, chen2020simple}. It explores different ways of sampling positives/negatives for each anchor and assumes positive pairs should have similar representations in the feature space. For example, \cite{He_2020_CVPR} uses a dictionary to maintain a negative set to increase the training variety. Bachman \textit{et al.}\cite{NEURIPS2019_ddf35421} use local features of two homologous images with different cropping as positive sample pair to help extract the details in images. 
However, features learned by CL generally capture instance-aware characteristics. To further facilitate the downstream classification task, it has been extended to supervised CL recently, where all images from the same class as an anchor are treated as positives and the rest as negatives \cite{khosla2020supervised}.  Some methods use self-supervised losses as auxiliary losses during supervised training \cite{gidaris2019boosting} or before supervised fine-tuning \cite{doersch2020crosstransformers}. The most related \cite{gao2021contrastive} further combines a supervised contrastive prototype learning with augmented embeddings based on a meta-learning training strategy for FSL. 
However, ours has vital differences. For each anchor, we take all samples in a batch into  the contrastive learning process and compare them with structural similarity, rather than only using samples from the same class as positives. In this way, both intra-class and inter-class visual variance can be simultaneously captured. And we use the transfer-learning strategy to learn generalized feature space, without any information from novel classes.

\section{Proposed Methods}
To solve few-shot learning (FSL) problem, we first introduce our decoupled learning strategy of feature embeddings and classifiers. Then, we detailedly present our dual-path feature learning architecture, that incorporates global structural similarity within contrastive embedding process. At last, we give a theoretical discussion about that the proposed method can learn a generalized feature space with increased inner-class alignment and inter-class uniformity. 

\subsection{Problem Definition}
Let us start by defining some notations. We follow the typical FSL setting. In FSL, we are given an abundant training set $\Cal{D}_b =\{(\V{x}_i, y_i) | y_i \in \Cal{C}_b, i = 1, ..., N_b\}$ from $|\Cal{C}_b|$ base classes, where $\V{x}_i$ is the $i$-th image, $y_i$ is the corresponding label and $N_b$ is the number of images in $\Cal{D}_b$. In addition, we also have a few-shot support set $\Cal{D}_n = \{(\V{x}_j, y_j) | y_j \in \Cal{C}_n, j = 1, ..., N\times K\}$ from $N$ novel classes (i.e. an $N$-way $K$-shot task), where each class has only $K$ (usually $K \leq 20$) labeled samples  and $\Cal{C}_b \cap \Cal{C}_n = \emptyset$. By exploiting $\Cal{D}_b$ and $\Cal{D}_n$ for training, FSL aims to learn a few-shot classifier for novel classes $\Cal{C}_n$. 

\subsection{Decoupled Learning of Feature Embeddings and Classifiers}
Our few-shot learning framework mainly includes two stages: the feature embedding stage and the classifier learning stage. 

\textbf{Feature embedding stage.} Learning a good generalized feature embedding space is one effective solution to the FSL problem \cite{tian2020rethinking}. It requires both inner-class alignment and inter-class uniformity.  Different from designing new meta learning based algorithms to learn the embedding space, we propose that an feature extractor  trained totally on base classes can also generate powerful embeddings for the downstream FSL novel tasks. To this end, we propose a dual-path strategy that fully exploits both class semantics and training images in base classes by incorporating global structural similarities, in order to learn a generalized and discriminative feature space. Once the feature embedding model $f_{\phi}$ is learned in the training process, we freeze it in the following test stage.

\textbf{Classifier learning stage.} In the test stage, we only need to learn a classifier upon the fixed embedding model with the support data in novel classes. We use the very simple prototype based classifier for each class, where each class prototype is computed as the mean vector of all support samples from the same class, i.e. 
\begin{equation}
	\V{p}_c = \frac{1}{K} \sum_{\forall \V{x}_i \in \Cal{D}_n, y_i = c} f_{\phi}(\V{x}_i), c \in \Cal{C}_b \cup \Cal{C}_n, 
	\label{eq:cls-prototype}
\end{equation}
where $\V{p}_c$ represents the prototype for class $c$. In order to evaluate efficacy of the above embedding model, we adopt both inductive and transductive inference for prediction. In \textit{inductive inference}, we calculate the cosine similarities between each test sample $\V{x}$ and all class prototypes and then predict its class label as following: 
\begin{equation}
	y = \arg \max_c \frac{\exp(cos(f_{\phi}(\V{x}), \V{p}_c))}{\sum_j \exp(cos(f_{\phi}(\V{x}), \V{p}_j))}. 
	\label{eq:cls-inductive}
\end{equation}
While in \textit{transductive inference}, we utilize all test samples to rectify class prototypes and reduce the possible model bias. We use Eq.\ref{eq:cls-prototype} and Eq.\ref{eq:cls-inductive} to compute the prediction scores of each test sample. Then we compute the rectified class prototype as the weighted sum of all test samples and the original prototype, defined as follows. 
\begin{equation}
	\tilde{\V{p}}_c = \frac{K\V{p}_c+\sum_{\forall \V{x} \in \Cal{D}_t} p(c|\V{x})f_{\phi}(\V{x})}{K+\sum_{\forall \V{x} \in \Cal{Q}} p(c|\V{x})}, 
	\label{eq:cls-prototype-rec}
\end{equation}
where $\Cal{D}_t$ denotes the test set. At last, we substitute the original prototype $\V{p}_c$ with $\tilde{\V{p}}_c$ in Eq.\ref{eq:cls-inductive} for the final prediction.

\subsection{Structure-Aware Contrastive Feature Learning}
In the feature embedding stage, we propose a dual-path learning architecture that are trained from only base samples via direct transfer learning strategy, as is shown in Fig.\ref{fig: dualpipe}. One path (i.e. base-path) predicts the structural similarity for each input and is then used as structure teaching for feature embedding in another path (i.e. novel-path). While, the latter combines the received similarity guidance with contrastive embedding effectively, in order to learn a generalized feature space for FSL.

\textbf{Extracting structural similarity matrix via pre-trained classifier. }
In the first step, we train a deep classifier $g_{\psi}$ in the standard manner with the training samples in all base classes. Each image in $\Cal{D}_b$ is randomly augmented $k$ times to synthesize multiple homologous images. For simplicity, we set $k=2$ in this paper,  where two homologous images of $\V{x}_i$ are denoted as $\tilde{\V{x}}_{2i-1}$ and $\tilde{\V{x}}_{2i}$ in the augmented dataset $\tilde{\Cal{D}}_b = {\{(\tilde{\V{x}}_n, \tilde{y}_n)\}}_{n=1}^{2N_b}, \tilde{y}_{2i-1}=\tilde{y}_{2i}=y_i$. We use $g_{\psi}$ to extract the structural similarity matrix $\V{S}_b$ for $\tilde{\Cal{D}}_b$ as follows: 
\begin{equation}
	\V{S}_b = \begin{bmatrix}
		\text{Softmax}(g_\psi(\tilde{\V{x}}_1)/{\tau_{hot}})^T\\
		\vdots  \\
		\text{Softmax}(g_\psi(\tilde{\V{x}}_{2N_b})/{\tau_{hot}})^T\\
	\end{bmatrix}
	\in \Set{R}^{2N_b \times C_b}
	\label{eq:structure-similarity}
\end{equation}
$\tau_{hot} \in \Set{R}^{+}$ is the temperature scalar, usually $\tau_{hot}>1$.  For any $\tilde{\V{x}}_n$, $S_b(n,c) = p(y=c|\tilde{\V{x}}_n)$ represents the posterior probability that it belongs to class $c$, or the similarity with class $c$. The use of temperature scalars is based on the idea of knowledge distillation \cite{hinton2015distilling}. Using predicting results divided by $\tau_{hot}$ before $softmax$ can make the distribution of prediction results more continuous and dispersed among classes. It larges the entropy of $logits$,  so as to increase the information in each class.

\textbf{Learning representation via structure-aware contrastive loss. } In the second step, we exploit the structural similarity between each homologous image and class to contrastively learn a generalized feature space in another path.  We propose a structure-aware contrastive loss (SACL). Different from the conventional un-/self-supervised CL loss or supervised CL loss (SCL), our SACL do not simply use homologous information or class labels in each training batch to sample positives/negatives for each anchor image. Instead, we treat all input images in a batch as positive-like samples, whose representation similarity is taught by the base path. In this way, the learned feature space can well balance the inter-class generalization ability and intra-class discrimination ability. 

Assume the homologous images of $\tilde{\V{x}}_i$ are indexed by $h(i)$, whose labels are $\tilde{y}_i$ and $\tilde{y}_{h(i)}$, respectively. We define the structure-aware contrastive loss as follows. 
\begin{equation}
	L_{sacl} = \sum_{\tilde{\V{x}}_i \in \Cal{D}_b} (- \sum_{j \in \Set{A}(i)} \frac{w_{ij}}{\sum_{j' \in \Set{A}(i)} w_{ij'}} L_{ij}),
	\label{eq:loss-sacl-total}
\end{equation}
where $L_{ij}$ is the loss between image $i$ and $j$ and $w_{ij}$ is the taught similarity, defined as: 
\begin{equation}
	\begin{split}
		& L_{ij} = \log \left[ \frac{\exp(\langle \V{e}_i, \V{e}_j\rangle/\tau_{cold})}{\sum_{j' \in \Set{A}(i)} \exp(\langle \V{e}_i, \V{e}_{j'}\rangle/\tau_{cold})}	
		\right] \\
		& w_{ij} = 	\begin{cases}
			1.0 &\text{if } j = h(i)\\
			\lambda*S_b(j,\tilde{y}_i) &\text{else}
		\end{cases}
	\end{split}
	\label{eq:loss-sacl-weight}
\end{equation}
$\V{e}_i = f_{\phi}(\tilde{\V{x}}_i) / ||f_{\phi}(\tilde{\V{x}}_i)||$ is the normalized feature embedding. $\langle \cdot \rangle$ denotes the inner product. $\Set{A}(i) = \Set{I} \backslash i $ are indices of all images but $i$. $\tau_{cold} \in \Set{R}^{+}$ is the temperature, generally less than 1. In the calculation of structure-aware contrastive loss, we divide the cosine similarity between two feature by $\tau_{cold}$ to make the output of $softmax$ more centralized, so that the feature extractor can focus on the most similar visual features between two images. In the training process, considering that each training batch is randomly generated, the performance of the base-path consequently varies. If a batch are predicted more decisively, it should play a bigger role in the feature learning process and vice versa. Taking this into consideration, we set $\lambda$ in Eq.\ref{eq:loss-sacl-weight} in an adaptive manner. We use the accuracy of $g_{\psi}$ in the batch for $\lambda$ in this paper. Training process is sketched in Alg.\ref{alg: SACL}.

\begin{algorithm}[tb]
	\caption{Training strategy of our proposed dual path structural contrastive embeddings. }
	\label{alg: SACL}
	\begin{algorithmic}
		\STATE {\bfseries Requires:} Base dataset $\Cal{D}_b$ which has $|\Cal{C}_b|$ classes, well-trained classifier $g_\psi$, feature extractor $f_\phi$ which outputs a $d$-dimension feature, temperatures $\tau_{hot}$ and $\tau_{cold}$.
		\STATE {\bfseries Begin:}
		\FOR{$iter=1$ {\bfseries to} $MaxIteration$}
		\STATE Randomly sample miniBatch $\mathcal{B}={\{\V{x}_n, y_n\}}_{n=1}^N$ from $\Cal{D}_b$
		\STATE Get augmented batch $\mathcal{A}={\{\tilde{\V{x}}_n, \tilde{y}_n\}}_{n=1}^{2N}$ by using random data augmentation on $\mathcal{B}$
		\STATE Compute structural similarity matrix $S_b$ with $g_{\psi}$
		\STATE Get features matrix $E=f_\phi(\mathcal{A})$, $E{\in}\mathbb{R}^{2N{\times}d}$
		\STATE Compute $L_{SACL}$
		\STATE Compute $\triangledown_{f_\phi}L_{SACL}$
		\STATE Update $f_\phi$ with $\triangledown_{f_\phi}$ use Adam optimization
		\ENDFOR
		\STATE {\bfseries return} Well trained feature extractor $f_\phi$.
	\end{algorithmic}
\end{algorithm}

\section{Analysis on Structure-Aware Contrastive Loss}

\subsection{Connection with Other Contrastive Loss}
We further give an analysis on our proposed SACL in the feature learning stage. For simplicity, we regard images with the same homology or the same category of an anchor in a batch as positives and the rest as negatives. We denote the structural similarities of positive samples for each anchor $\V{x}_i$ as $\Set{P}(i)$ and the negative samples as $\Set{N}(i)$, which can be achieved by a well trained deep classifier. All proof procedures in this section can be found in the appendix \ref{apx: Theory Analyze}.

First, Eq.\ref{eq:loss-sacl-total} boils down to the popular self-supervised contrastive loss in the degenerated case where the normalized weight $\tilde{w}_{ij} \approx 1$ for all $y_j = y_i$ and $\tilde{w}_{ij} \approx 0$ otherwise. This happens under the condition of a \textit{near perfect} classifier $g_{\psi}$, which produces an approximately binary similarity matrix $\V{S}_b$. When $\Set{P}(i) = \{j: y_j =y_i\}$ is a singleton (i.e. $j = h(i)$), SACL reduces to CL. When $\Set{P}(i)$ contains multiple samples of the same class, it reduces to SCL. 

Second, we show that under certain assumptions, asymptotically $L_{sacl}$ converges to a combination of alignment and uniformity loss proposed in \cite{pmlr-v119-wang20k}. 
\begin{definition}[Consistency assumption] Assume $i$-th sample is denoted by $\V{x}_i$. 
	Let $n_{y_i} = |\{j: y_j=y_i\}|$. We say the classifier in SACL is consistent if for a new sample $\V{x}_j$, 
	$$w_{ij} = \begin{cases}
		1-o(\frac{1}{n_{y_i}}) & \text{if } y_j = y_i \\
		o(\frac{1}{n_{y_i}}) & \text{if } y_j \neq y_i \\
	\end{cases}
	$$
	where $\lim_{n_{y_i} \rightarrow \infty} n_{y_i}o(\frac{1}{n_{y_i}})=0$. In other words, we are able to classify class $y_i$ more accurate as sample size $n_{y_i}$ grows large, and the order of deviation is linear, i.e. $\frac{1}{n_{y_i}}$.
\end{definition}
Thus, if a deterministic decision boundary exists and it can be approximated by neural networks at any precision, then the oracle bound is attained when sample size goes to infinity.

\begin{theorem}[SACL as alignment and uniformity loss]\label{thm:WCL_align_uni} 
	Suppose there are $n$ samples in total from $K$ classes and each class has $n_k$ samples, so that  $n=\sum_{k=1}^K n_k$. Also suppose that no single class dominates, i.e. $\lim_{n_k,n\rightarrow\infty}\frac{n_k}{n}=p_k>0,\ k=1,\cdots,K.$ Then under the consistency assumption, we have: 
	
	(1) As $n, n_k$ goes to infinity, $	\lim_{n\rightarrow\infty}\left(L_i-\log n\right)=-\mathbb{E}_{a\sim p_{pos}}\langle \V{e}_i,\V{e}_a \rangle_\tau+\log\mathbb{E}_{a\sim p_{data}} \exp(\langle \V{e}_i,\V{e}_a \rangle_\tau)$. $p_{pos}$ is the distribution of positive samples and $p_{data}$ describes the whole distribution. The first term is known as \textbf{alignment loss} and the second is \textbf{uniformity loss}, according to \cite{pmlr-v119-wang20k}. 
	
	(2) $L_{sacl}$ is a weighted sum of leave-one-out (LOO) alignment and uniformity loss.
	
	(3) The rate of convergence of $L_i-\log n$ to the right-hand-side is $O(\exp(-Cn))$ where $C$ is a universal constant.	
\end{theorem}

The first part of the theorem tells us that for each $\V{x}_i$, $L_i-\log n$ can be approximately decomposed into two terms:
$$L_i-\log n\approx \text{ alignment loss }+\text{ uniformity loss}$$
Minimizing the above loss means aligning each sample $\V{x}_i$ with its positive counterparts while pulling away from its negatives. It benefits FSL for the following possible reasons: (1) Assume we have three base classes, e.g. \textit{dog, cat, rooster}, and one novel class, \textit{duck}. Intuitively, ``duck'' should be close to ``rooster'' and way from the other two classes in the extracted feature space. ``Close" refers to the alignment loss because the sample distribution of \textit{duck} and \textit{rooster} are similar and``away from" refers to the uniformity loss. (2) The second part suggests that we can replace $\V{x}_i$ with a subset of samples from $\mathbb{I}$ and such modification corresponds to bootstrapping in statistics. (3) The last part states that the convergence is exponentially fast and is independent of the number of classes $K$.

\subsection{Backward Propagation Analysis}

In this subsection, we prove that classifier and cold temperature are helpful for enlarging the effects of hard positive in backward propagation as well as making model focus on visual similar images. We give the final derivation formula of hard positive samples on the backward propagation of gradient of $L^{\text{SACL}}$ with respect to $f_\phi()$ as follows:

\begin{align}
	\label{equ: hardpos}
	&k\sum_{a{\in}\mathbb{P}(i)}\exp(\langle \V{e}_i,\V{e}_a\rangle_{\tau})+k\sum_{a{\in}\mathbb{N}(i)}\exp(\langle \V{e}_i,\V{e}_a\rangle_{\tau})\nonumber\\
	&-(\|\mathbb{P}(i)\|k+\|\mathbb{N}(i)\|)
\end{align}

Considering that easy positive and hard negative samples have similar visual concepts with anchor image, we assume $\langle \V{e}_i,\V{e}_a\rangle>0$ for $a\in\mathbb{P}(i)\cup\mathbb{N}(i)$. With the increasing of $k$ and the decreasing of $\tau$, the value of Eq.\ref{equ: hardpos} will become larger, which means the effects of hard positive in gradient descent becomes larger. Therefore, when anchor image compares with hard positive, the structural similarity and cold temperature can help model learn more from visual similar images (easy positive and hard negative) as our original purpose of proposing SACL. More details can be found in appendix \ref{apx: backward analyze}.

\begin{table*}[t]
	\centering
	\caption{Comparison with state-of-the-arts on \textit{mini}ImageNet and CUB datasets.  Average accuracy with 95\% confidence intervals are reported. The top/bottom part includes inductive/transductive methods. }
	\label{table: fsl-mini}
	\begin{small}
		\begin{tabular}{|l|c|cc|cc|}
			\hline
			\multirow{2}{*}{Method}	&	\multirow{2}{*}{Backbone}	&	\multicolumn{2}{c|}{miniImageNet}	&	\multicolumn{2}{c|}{CUB} \\ 
			\cline{3-6} 
			&		&	5-way 1-shot	&	5-way 5-shot	&	5-way 1-shot	&	5-way 5-shot \\ 
			\hline
			
			MAML \cite{finn2017modelagnostic}	&	Conv4	&	48.70 $\pm$ 0.84	&	55.31 $\pm$ 0.73	&	54.73 $\pm$ 0.97	&	75.75 $\pm$ 0.76 \\
			MatchingNet \cite{NIPS2016_90e13578}	&	Conv4	&	43.44 $\pm$ 0.77	&	60.60 $\pm$ 0.71	&	60.52 $\pm$ 0.88	&	75.29 $\pm$ 0.75 \\
			ProtoNet \cite{snell2017prototypical}	&	Conv4	&	49.42 $\pm$ 0.78	&	68.20 $\pm$ 0.66	&	50.46 $\pm$ 0.88	&	76.39 $\pm$ 0.64 \\
			RelationNet \cite{Sung_2018_CVPR}	&	Conv4	&	50.44 $\pm$ 0.82	&	65.32 $\pm$ 0.70	&	62.34 $\pm$ 0.94	&	77.84 $\pm$ 0.68 \\
			DN4 \cite{Li_2019_CVPR}	&	Conv4	&	51.24 $\pm$ 0.74	&	71.02 $\pm$ 0.64	&	53.15 $\pm$ 0.84	&	81.90 $\pm$ 0.60 \\
			GCR \cite{li2019few}						&	Conv4	&	53.21 $\pm$ 0.40	&	72.34 $\pm$ 0.32	&	-	&	-\\
			DSN \cite{simon2020adaptive}	&	Conv4	&	51.78 $\pm$ 0.96	&	68.99 $\pm$ 0.69 &	-	&	-\\
			Baseline++ \cite{chen2020closer}	&	Conv4	&	48.24 $\pm$ 0.75	&	66.43 $\pm$ 0.63	&	60.53 $\pm$ 0.83	&	79.34 $\pm$ 0.61 \\
			FEAT \cite{ye2020fewshot}&  Conv4 & 55.15 $\pm$ 0.20 & 71.61 $\pm$ 0.16 & - & - \\
			\textbf{Ours}	&	Conv4	&	\textbf{58.27 $\pm$ 0.66}	&	\textbf{74.39 $\pm$ 0.47}	&	\textbf{67.39 $\pm$ 0.73}	&	\textbf{82.59 $\pm$ 0.43} \\
			MTL \cite{Sun_2019_CVPR}	&	ResNet-12	&	61.20$\pm$1.80	&	75.50$\pm$0.80	&	-	&	- \\
			LEO \cite{rusu2019metalearning}	&	WRN-28	&	61.76$\pm$0.08	&	77.59$\pm$0.12	&	-	&	- \\
			MetaOpt \cite{Lee_2019_CVPR} &	ResNet-12	&	62.64$\pm$0.62	&	78.63$\pm$0.46	&	-	&	- \\
			Meta-Baseline \cite{chen2020new}	&	ResNet-12	&	63.17$\pm$0.23	&	79.26$\pm$0.17	&	-	&	- \\
			TADAM \cite{oreshkin2018tadam}	&	ResNet-12	&	58.50$\pm$0.30	&	76.70$\pm$0.30	&	-	&	- \\
			CAN \cite{NEURIPS2019_01894d6f}	&	ResNet-12	&	63.85$\pm$0.48	&	79.44$\pm$0.34	&	-	&	- \\
			FEAT \cite{ye2020fewshot}	&	ResNet-12	&	\underline{66.78 $\pm$ 0.20}	&	\textbf{82.05 $\pm$ 0.14}	&	-	&	- \\
			RFS-Distill \cite{tian2020rethinking} & ResNet-12 & 64.82 $\pm$ 0.60 & 82.14 $\pm$ 0.43 & - & - \\			
			\textbf{Ours}	&	ResNet-12	&	\textbf{66.87$\pm$0.66}	&	\underline{81.57$\pm$0.46}	&	\textbf{77.20$\pm$0.66}	&	\textbf{89.23$\pm$0.33} \\
			\hline\hline
			TPN \cite{liu2019learning}	&	ResNet-12	&	59.49	&	75.65	&	-	&	- \\
			CAN \cite{NEURIPS2019_01894d6f}	&	ResNet-12	&	67.19$\pm$0.55	&	80.64$\pm$0.35	&	-	&	- \\
			Trans fine-tune \cite{dhillon2019baseline}	&	WRN-28-10	&	68.11$\pm$0.69	&	80.36$\pm$0.50	&	-	&	- \\
			LaplacianShot \cite{pmlr-v119-ziko20a}	&	ResNet-18	& 72.11$\pm$0.19	&	 82.31$\pm$0.14	& 80.96	&  88.68 \\
			FEAT \cite{ye2020fewshot}	&	Conv4	&	57.04$\pm$0.20	&	72.89$\pm$0.16	&	-	&	- \\
			\textbf{Ours}	&	Conv4	&	64.73$\pm$0.88	&	75.93$\pm$0.53	&	76.26$\pm$0.91	&	83.47$\pm$0.44 \\
			\textbf{Ours}	&	ResNet-12	&	\textbf{75.35$\pm$0.81}	&	\textbf{83.53$\pm$0.45}	&	\textbf{86.52$\pm$0.68}	&	\textbf{90.26$\pm$0.33} \\
			\hline
		\end{tabular}
	\end{small}
\end{table*}

\section{Experiments}
In this section, we evaluate our approach by conducting three groups of experiments: (1) standard FSL setting where the label search space is restricted to only novel classes, (2) generalized FSL setting where the label space includes both base and novel classes, and (3) ablation studies. 
\subsection{Experimental Setups}
\subsubsection{Datasets}
We evaluate our method on three popular benchmark FSL datasets: \textit{mini}ImageNet \cite{NIPS2016_90e13578}, \textit{tiered}ImageNet \cite{ren2018metalearning} and Caltech-UCSD Birds-200-2011 (CUB) \cite{WahCUB_200_2011}. \textit{mini}ImageNet contains 100 classes randomly selected from ImageNet and each class have 600 images with resolution of $84\times 84$. We follow the standard dataset split in prior works and uses 64 classes for training, 16 classes for validation and 20 classes for testing. \textit{tieredImageNet} is a larger subset of ImageNet. It contains 608 classes from 34 super categories and 779,165 images in total. Each image is of size $224 \times 224$ and resized to $84 \times 84$ in the experiments. We take 351 classes, 97 classes and 160 classes for training,  validation and testing, respectively. CUB is a fine-grained dataset, containing 200 bird species and 11,788 images in total. Following the setup in prior works, we take 100, 50, 50 classes for training, validation and testing, respectively. 

\subsubsection{Implementation details} 
We implement all of our experiments in Pytorch  with a NVIDIA 1080Ti GPU and use Adam \cite{kingma2014adam} to optimize the whole network end-to-end over 200 epochs. For fair comparison, we use both Conv4 \cite{chen2020closer}  and ResNet-12 \cite{NEURIPS2019_01894d6f} as the network backbone and a fc layer on top to form $g_{\psi}$. For Conv4, we use a linear projector as in contrastive learning \cite{chen2020simple}. In the training stage, we adopt cropping, color jittering and random grayscale conversion to augment training samples. We set the learning rate in Adam to $10^{-3}$, $10^{-3}$ and $2\times10^{-4}$ for \textit{mini}ImageNet, \textit{tiered}ImageNet and CUB, respectively.  $\tau_{hot}$ is 2.5 and $\tau_{cold}$ is 0.05 for all benchmarks.

\paragraph{Evaluation protocols} 
Following the standard FSL setting adopted by most FSL works \cite{rusu2019metalearning}, we conduct 5-way 1-shot and 5-way 5-shot classification on benchmark datasets. Each class has 1 or 5 samples for training and another 15 random samples for evaluation/testing. We construct 1,000 episodes and report the average precision of those tasks with the 95\% confidence interval, to measure the effectiveness of the proposed method. 

\subsection{Experimental Results on Standard FSL Setting}
In the first set of experiments, we evaluate the proposed method on its performance under the standard FSL setting, where test samples are from novel classes and the label search space is also restricted to only novel classes.  Table.\ref{table: fsl-mini} and Table.\ref{tab:fsl-tiered} provide comparative results on all three benchmarks. From these results, we can observe: (1) The proposed method can achieve the best on almost all tasks when using different feature backbone.  The independence from different backbones can validate the effectiveness of our dual path structural contrastive learning, to some extent. (2) With either inductive or transductive inference, our approach can obtain the best results. For example, when using Conv4 as backbone, ours has a significant performance gain on three datasets than other FSL methods. For 5-way 1-shot tasks on \textit{mini}ImageNet, it achieves more than 3\% ahead of the second best method. This indicates that in severe data-scarce situation, using structural similarity in contrastive feature construction can learn a more generalized feature space for FSL. 

Our method is slightly worse than FEAT and CAN in some cases. For example, when using ResNet-12 as backbone, ours achieves 0.5\% less than FEAT for 5-way 5-shot tasks on \textit{mini}ImageNet. However, comparing with 55,041KB parameters in FEAT, ours has 31,454KB. Even with less few parameters, it can still achieve comparable performance. While comparing with other baselines \cite{finn2017modelagnostic, snell2017prototypical, NIPS2016_90e13578}, the base path in our method is only pre-trained once. It won't consume too much training sources, but can improve the performance a lot.  

\begin{table}[!htbp]
	\centering
	\caption{Performance comparison with other methods on 5-way 1-shot and 5-way 5-shot classification tasks on \textit{tiered}ImageNet dataset. Average accuracy with 95\% confidence intervals are reported. $\dagger$ represents using transductive inference in test stage.}
	\label{tab:fsl-tiered}
	\begin{tabular}{lccc}
		\hline
		Methods     & Backbone    & 5-way 1-Shot  &  5-way 5-Shot \\
		\hline
		MAML 	&	Conv4	&	51.67$\pm$1.81	&	70.30$\pm$1.75 \\
		ProtoNet	&	Conv4	&	53.31$\pm$0.89	&	72.69$\pm$0.74 \\
		RelationNet&	Conv4	&	54.48$\pm$0.93	&	71.32$\pm$0.78 \\
		\textbf{Ours}	&	Conv4	&	58.80$\pm$0.69	&	74.84$\pm$0.59 \\
		\textbf{Ours$\dagger$}	&	Conv4	&	\textbf{66.63$\pm$0.92}	&	\textbf{76.61$\pm$0.60} \\
		\hline
		DSN &	ResNet-12	&	66.22$\pm$0.75	&	82.79$\pm$0.48 \\
		MetaOpt &	ResNet-12	&	65.99$\pm$0.72	&	81.56$\pm$0.53  \\
		LEO &	WRN-28	&	66.33$\pm$0.05	&	81.44$\pm$0.09  \\
		FEAT 	&	ResNet-12	&	\textbf{70.80 $\pm$0.23}	&	\textbf{84.79$\pm$0.16} \\		
		CAN 	&	ResNet-12	&	69.89$\pm$0.51	&	84.23$\pm$0.37 \\
		CAN$\dagger$ 	&	ResNet-12	&	73.21$\pm$0.58	&	\textbf{84.93$\pm$0.38} \\
		\textbf{Ours}	&	ResNet-12	&	69.23$\pm$0.76	&	82.62$\pm$0.56 \\
		\textbf{Ours$\dagger$}	&	ResNet-12	&	\textbf{77.24$\pm$0.88}	&	84.41$\pm$0.53 \\
		\hline
	\end{tabular}
\end{table}

\begin{table}[h]
	\centering
	\caption{Performance comparison on \textit{mini}ImageNet for 100-way tasks under gFSL setting. $acc_b$/$acc_n$ denotes the accuracy of classifying test samples from base/novel classes to all classes. $acc_t$ is the accuracy of classifying all test samples to all the classes. $\ddag$ represents training on full base training samples, rather than the same subset as others. }
	\label{table:gfsl-mini}
	\begin{tabular}{lcccc}
		\hline
		Methods   &  Backbone    & $acc_{t}$  &  $acc_b$	&$acc_n$\\
		\hline
		MatchingNet	&	Conv4	&	26.98	&	33.54	&	0.75 \\
		ProtoNet	&	Conv4	&	31.17	&	39.53	&	0.52 \\
		RelationNet &	Conv4	&	32.48	&	40.24	&	1.42 \\
		GCR 	&	Conv4	&	39.14	&	46.32	&	12.98 \\
		FEAT $\ddag$			&	Conv4	&	35.48	&	40.16	&	16.75 \\
		\textbf{Ours}					&	Conv4	&	\textbf{49.98}	&	\textbf{56.14}	&	\textbf{25.35} \\
		\hline
		FEAT $\ddag$			&	ResNet-12	&	70.64	&	80.65	&	30.43 \\
		\textbf{Ours} $\ddag$					&	ResNet-12	&	\textbf{74.65}	&	\textbf{85.00}	&	\textbf{33.25} \\
		\hline
	\end{tabular}
\end{table}

\subsection{Experimental Results on Generalized FSL Setting}
In the second set of experiments, we evaluate the proposed method on its performance under the generalized FSL setting. In this setting, test samples are selected from both base classes and novel classes and their labels are predicted from the joint label space of base classes and novel ones. Comparing with FSL, it is more challenging and realistic, since we cannot know which domain an incoming sample is from beforehand. We follow the setting in \cite{li2019few} to conduct a 100-way generalized few-shot test on  \textbf{\textit{mini}ImageNet}. Concretely, classes in \textit{mini}ImageNet are split into 64/16/20 as training/validation/test with 600 images in each class. We randomly select 500 images from each class in training and validation sets to construct the training base classes.  Then we select the rest 100 images from each class in training and validation sets to form test base classes, and randomly select 100 images from each class in test set to form test novel classes. The training strategy is the same as FSL, except that the model is trained on new training base classes.
\begin{figure}[!htbp]
	\centering
	\includegraphics[width=0.4\textwidth, height=0.22\textheight]{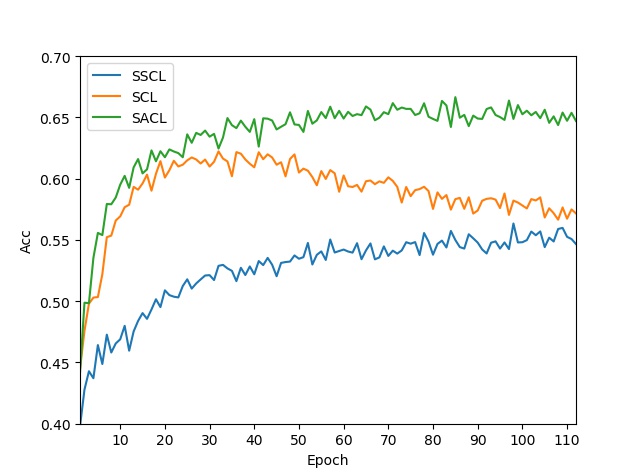}
	\caption{The 5-way 1-shot accuracy trend on miniImageNet during training.}
	\label{fig: mini_trend}
\end{figure}

We evaluate all methods using three metrics, i.e. $acc_b$, $acc_n$ and $acc_t$.  $acc_b$ ($acc_n$) denote  the classification accuracy of base (or novel) samples with the label search space being the joint space. $acc_t$ is their harmonic mean, i.e. $acc_t = \frac{2* acc_b * acc_n}{acc_b + acc_n}$. It describes the ability of balancing between base and novel domains. We choose the harmonic mean as main criterion to favor high accuracies on both base and novel classes. We compare our proposed method with other five few-shot learning methods on 100-way GFSL task. The results of MatchingNet \cite{NIPS2016_90e13578}, ProtoNet \cite{snell2017prototypical}, RelationNet \cite{Sung_2018_CVPR} and GCR \cite{li2019few} have the same experimental setting as our method. The model parameters of FEAT \cite{ye2020fewshot} is downloaded from github link offered by \cite{ye2020fewshot}. It trains and validates the model on full base classes rather than the new base classes. As shown in Tab.\ref{table:gfsl-mini}, ours has a significant better performance on 100-way gFSL task.

\begin{table}[!htbp]
	\centering
	\caption{Performance comparison in FSL. We report the 5-way classification accuracy with 95\% confidence interval. $\dagger$ represents using transductive inference in test stage. We use ResNet-12 as the backbone.}
	\label{table: ablation-struc}
		\begin{tabular}{l cccc}
			\hline
			\multirow{2}{*}{Method}	&	\multicolumn{2}{c}{miniImageNet}	&	\multicolumn{2}{c}{CUB} \\ 
			\cline{2-5}
			&		1-shot	&	5-shot	&	1-shot	&	5-shot  \\
			\hline
			CL  & 55.28$\pm$ 0.66 & 71.76$\pm$ 0.52 & 37.20$\pm$ 0.49 & 51.53$\pm$ 0.54\\
			CL$\dagger$   & 61.74$\pm$ 0.86 & 73.42$\pm$ 0.55 & 38.98$\pm$ 0.59 & 52.71$\pm$ 0.57\\
			SCL    & 60.93$\pm$ 0.69 & 74.94$\pm$ 0.52 & 74.82$\pm$ 0.69 & 87.07$\pm$ 0.38 \\
			SCL$\dagger$   & 67.24$\pm$ 0.82 & 76.27$\pm$ 0.51 & 82.10$\pm$ 0.77 & 88.17$\pm$ 0.38 \\
			SACL    & 66.87$\pm$ 0.66 & 81.57$\pm$ 0.46 & 77.20$\pm$ 0.66 & 89.23$\pm$ 0.33\\
			SACL$\dagger$   & \textbf{75.35$\pm$ 0.81} & \textbf{83.53$\pm$ 0.45} & \textbf{86.52$\pm$ 0.68} & \textbf{90.26$\pm$ 0.33} \\
			\hline
		\end{tabular}
\end{table}
\begin{figure*}[!htp]
	\centering
	\includegraphics[width=\textwidth]{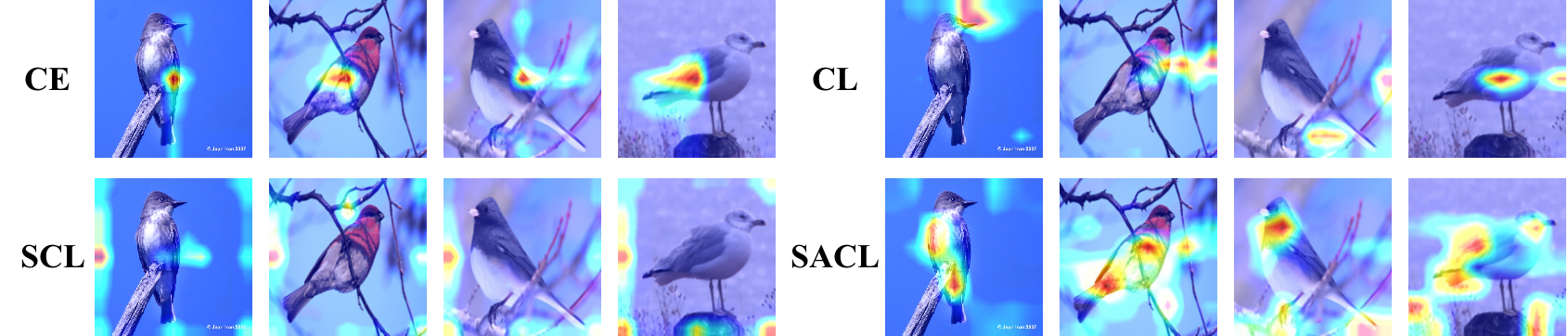}
	\caption{Visualization of high responding regions in CUB by using different losses in feature learning.  CE, CL, SCL and SACL are cross-entropy, self-supervised contrastive,  supervised contrastive and our structure-aware contrastive loss, respectively. SACL can preserve more information on novel classes, inducing better generalization ability. }
	\label{fig: visualize}
\end{figure*}

\subsection{Ablation Studies}
\subsubsection{Influence of structural similarity}
\begin{figure*}[!htp]
	\centering
	\includegraphics[width=\textwidth]{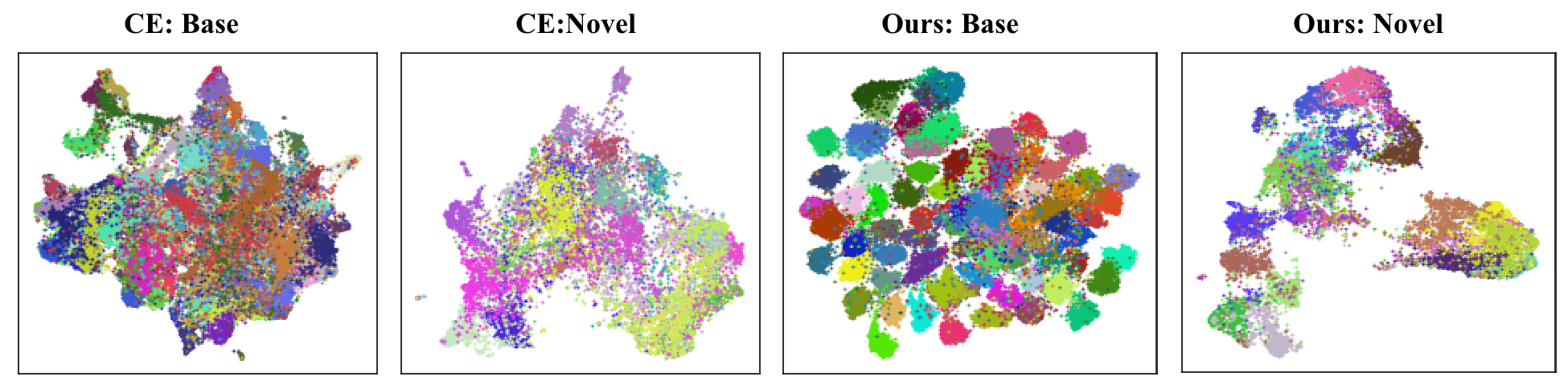}
	\caption{UMAP visualization \cite{mcinnes2018umap} of data embeddings from both base and novel classes, learned by cross-entropy loss and our SACL loss. Dots with different colors represent data points from different classes. Ours can well balance between the inter-class generalization and intra-class discrimination.}
	\label{fig: visualize-features}
\end{figure*}

\begin{figure}[t]
	\centering
	\includegraphics[width=0.4\textwidth, height=0.22\textheight]{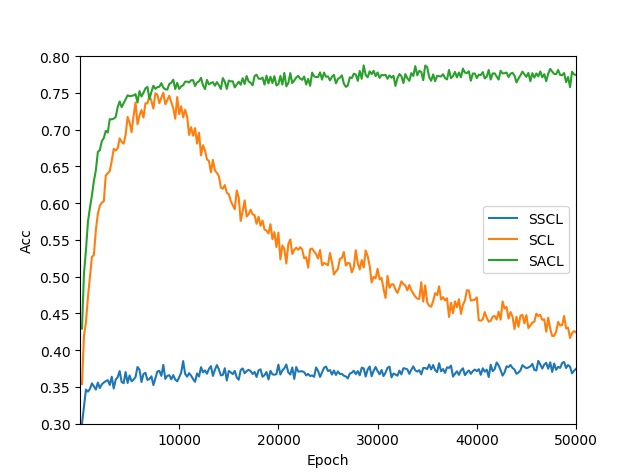}
	\caption{The 5-way 1-shot accuracy trend on CUB during training.}
	\label{fig: cub_trend}
\end{figure}

Structural similarity is a key factor affecting the feature distribution. It encourages the feature space to be more generalized, thus leading to a performance boost for FSL. We compare SACL with SCL using ground-truth one-hot labels as similarity and CL using none in Tab.\ref{table: ablation-struc}. SACL can significantly outperform others under both inductive and transductive inference. This validates the efficacy of structural similarity in feature space learning.  We also visualize the high responding regions in backward propagation of well-trained models under different losses, in order to figure out how SACL learns the similarity information. As shown in Fig.\ref{fig: visualize}, SACL focuses on multiple discriminative regions of novel objects, meaning that more visual details can be captured by our feature space across different classes. This also verifies the generalization ability of our proposed method. In addition, as shown in Fig.\ref{fig: mini_trend} and Fig.\ref{fig: cub_trend}, the use of structural similarity makes our method have better performance than CL and better convergence effect than SCL.

\subsubsection{Visualization of feature embeddings.} We plot the feature embeddings of both base and novel classes in low-dimension space with UMAP \cite{mcinnes2018umap} in Fig.\ref{fig: visualize-features}. As can be seen, the feature embeddings (either base or novel classes) learned by our SACL loss are more clustered within each class and less overlapped among different classes.

\subsubsection{Effect of temperatures}

In our proposed method, we use temperature $\tau_{hot}$ to adjust the smoothness of structural similarity distribution in Eq.\ref{eq:structure-similarity}.  From the results in Tab.\ref{table:temperature}, when $\tau_{hot}$ goes larger, the accuracy gradually decreases. According to our analysis in the supplementary material, the structural similarity would be evenly distributed, when  $\tau_{hot}$ gets large. This would decrease the multiple relationships between positive and negative samples, thus affecting the backward propagation. $\tau_{cold}$ is used to centralize the output of $softmax$ function, in order to highlight the influence of similar features in the final contrastive learning process.  A relative small value of $\tau_{cold}$ is more appropriate in FSL. 
\begin{table}[!htbp]
	\centering
	\caption{5-way 1-shot performances on \textit{mini}ImageNet with different $\tau_{cold}$ and $\tau_{hot}$. }
	\label{table:temperature}
	\begin{tabular}{|c| c c c |}
		\hline
		\multirow{2}{*}{$\tau_{cold}$}  &   \multicolumn{3}{c|}{$\tau_{hot}$} \\
		\cline{2-4}
		&	2.5	&	5.0	&	7.5 \\
		\hline
		
		0.05    & 53.94   & 52.51   & 51.27  \\
		
		0.10    & 53.57   & 52.69   & 50.13  \\
		
		0.50    & 47.90   & 47.93   & 46.46  \\
		\hline
	\end{tabular}
\end{table}

\subsubsection{Choice of batch size}
We also give 5-way 1-shot classification results on miniImageNet to observe the impact of different batch size on FSL.  When batch size increases from 64, 128, 256, 512 to 1024, the accuracy goes from 51.28\%, 52.69\%, 53.48\% to 52.01\%. Increasing the batch size properly is helpful to improve FSL performance. However, a larger batch size may harm the performance,  since it will increase the value of $(\|\mathbb{P}(i)\|k+\|\mathbb{N}(i)\|)$ in the back-propagation process.

\section{Conclusion}
We proposed a dual path contrastive feature learning scheme in the decoupled learning strategy for few-shot learning.  It combines the structural similarities with contrastive feature construction effectively, so that even without any information from novel classes, we could still learn generalized feature embeddings. Thus inner-class alignment and inter-class uniformity can be well guaranteed. Experimental results validate the its effectiveness and efficiency.


\bibliographystyle{ieee_fullname}
\bibliography{citation_TNNLS}

{\appendices
	
	\section{Proof of Theorem 1}
	\label{apx: Theory Analyze}
	We prove the Theorem using a probabilistic argument. SACL regards all other images in a batch as positives of an anchor image. For the convenience of later proof, we call images with the same label as the anchor $\V{x}_i$ as positives, which are indexed by  $\Set{P}(i) = \{j \in \Set{A}(i): y_j = y_i\}$. Images with different labels are called negatives, indexed by $\Set{N}(i) = \{ j \in \Set{A}(i): y_j \neq y_i\}$. We use $\tau$ to represent $\tau_{cold}$ for simplicity. $w_{ij}$ represents the similarity between $\tilde{\V{x}}_i$ and $\tilde{\V{x}}_j$ given by classifier, so similarity of positive pair $w_{ip}$ will be larger than that of negative pair $w_{in}$.  We assume $w_{in}=w$ ( a probability radix) and $ w_{ip}=kw$, where $k\in\mathbb{N}^+$ is a multiple relationship between $w_{in}$ and $w_{ip}$. 
	
	\subsection{Two results in probability theory}
	For the completeness of the proof, we first cite two classical results in probability theory. 
	
	\begin{theorem}[Law of large numbers]
		Suppose $X_i, i=1,2,\cdots n,\cdots$ is a sequence of i.i.d random variables with finite expectation $\mu=\mathbb{E}X_i$, then:
		\begin{itemize}
			\item $\lim_{n\rightarrow\infty}\mathcal{P}(|\frac{1}{n}\sum_{i=1}^nX_i-\mu|\ge\epsilon)=0$ for arbitrarily small positive $\epsilon$.
			\item $\lim_{n\rightarrow\infty}\frac{1}{n}\sum_{i=1}^nX_i=\mu$ almost surely.
		\end{itemize}
	\end{theorem}
	
	\begin{theorem}[General Hoeffding's inequality] Suppose $X_i, i=1,2,\cdots n,\cdots$ is a sequence of i.i.d random variables bounded within $[-1, 1]$, then we have
		$$\mathcal{P}(|\frac{1}{n}\sum_{i=1}^nX_i-\mu|\ge\epsilon)\le2\exp(-\frac{n\epsilon^2}{2})$$
		where $\mu=\mathbb{E}X_i$ and $\epsilon$ is an arbitrarily small positive number.
	\end{theorem}
	
	Note that because the statement (2) in Theorem 1 follows directly from statement (1), we shall only prove statement (1) and statement (3) in the main paper. 
	We shall only prove part one and the part three.
	\subsection{Proof of statement (1)}
	Decompose $L_i$ into two parts:
	\begin{align}
		L_i&=-\sum_{j \in \mathbb{P}(i)} \tilde{w}_{ij}L_{ij}-\sum_{j\in\mathbb{N}(i)}\tilde{w}_{ij}L_{ij}, 
	\end{align}
	we first show that the second summation is $o(1)$ and thus is negligible. By the consistency assumption and the fact that no single class dominates others, we have
	\begin{align*}
		\sum_{j \in\mathbb{A}(i)}w_{ij}& = \sum_{j\in\mathbb{P}(i)}w_{ij}+\sum_{j\in\mathbb{N}(i)}w_{ij}\\
		&=\left(n_k-1-o(1)\right) + o\left(n/n_k-1\right)\\
		&=n_k+o(1)
	\end{align*}
	The third equality follows from $o(1)+o(1)=o(1)$. Thus,
	
	\begin{align*}
		\widetilde{w}_{ij}&=\frac{w_{ij}}{\sum_{j\in\mathbb{A}(i)}w_{ij}}\\
		&=\begin{cases}
			\frac{1}{n_k}+o(\frac{1}{n_k^2})&\text{ if }j \in\mathbb{P}(i).\\
			o(\frac{1}{n_k^2})&\text{ if } j \in\mathbb{N}(i).
		\end{cases}
	\end{align*}
	where the second equality follows from $\frac{o(1)}{n_k}=o(\frac{1}{n_k})$ and $\frac{1}{n_k}\sim\frac{1}{n_k+o(1)}$.
	
	Next, rewrite $L_{ij}$ as
	\begin{align}
		L_{ij}=\langle \V{e}_i,\V{e}_j\rangle_\tau-\log\left(\sum_{j'\in\mathbb{A}(i)}\exp(\langle \V{e}_i,\V{e}_j' \rangle_\tau)\right)
	\end{align}
	
	We have 
	$$-\frac{2}{\tau}-\log n\leq L_{ij}\leq \frac{2}{\tau}-\log n$$. 
	
	The left inequality is attained when $\V{e}_i=\V{e}_j$ and $\V{e}_i=-\V{e}_j'\ \forall j'\in\mathbb{A}(i),j'\not=j$. While, the right inequality is attained when $\V{e}_i=-\V{e}_j$ and $\V{e}_i=\V{e}_j'\ \forall j'\in\mathbb{A}(i), j'\not=j$. Thus, combining with the previous results,
	
	\begin{equation}
		\begin{split}
			& o(\frac{1}{n_k^2}) \sum_{j\in\mathbb{N}(i)}\left(-\frac{2}{\tau}+\log n\right) \leq -\sum_{j\in\mathbb{N}(i)}\widetilde{w}_{ij}L_{ij} \\
			&\leq o(\frac{1}{n_k^2}) \sum_{j\in\mathbb{N}(i)}\left(\frac{2}{\tau}+\log n\right)
		\end{split}
	\end{equation}
	
	Both left and right hand side are bounded by
	$$o(\frac{1}{n_k^2}) O(n\log n)=o(\frac{\log n}{n_k})\le o(1)$$  Hence, we have shown
	\begin{align}
		-\sum_{a\in\mathbb{N}(i)}\widetilde{w}_{ia}L_{ia}&\sim o(1)
	\end{align}

	Now focus on the first summation of $L_i$:
	\begin{align*}
		& -\sum_{j \in\mathbb{P}(i)} \widetilde{w}_{ij}L_{ij}=-\sum_{j\in\mathbb{P}(i)}\frac{1}{n_k-1}\langle \V{e}_i,\V{e}_j\rangle_\tau\\ 
		& +\sum_{j\in\mathbb{P}(i)}\frac{1}{n_k-1}\log\left(\sum_{j'\in\mathbb{A}(i)}\exp(\langle \V{e}_i,\V{e}_j' \rangle_\tau)\right)\\
		& -o(\frac{1}{n_k^2})\sum_{a\in\mathbb{P}(i)}L_{ij}
	\end{align*}
	The term $o(\frac{1}{n_k^2})\sum_{j \in\mathbb{P}(i)}L_{ij}$ is $o(1)$ and hence negligible.
	
	By law of large numbers and the fact that $\langle \V{e}_i, \V{e}_j \rangle_\tau$ is bounded, we have
	$$\lim_{n_k \rightarrow\infty}\left(-\sum_{j\in\mathbb{P}(i)}\frac{1}{n_k}\langle \V{e}_i,\V{e}_j\rangle_\tau\right)=-\mathbb{E}_{j \sim p_{pos}}\langle \V{e}_i,\V{e}_j \rangle_\tau$$
	almost surely. 
	
	The middle term can be re-written as
	\begin{equation}
		\begin{split}
			& \sum_{j\in\mathbb{P}(i)}\frac{1}{n_k-1}\log\left(\sum_{j'\in\mathbb{A}(i)}\exp(\langle \V{e}_i,\V{e}_{j'} \rangle_\tau)\right) \\
			&  =\log\left(\frac{\sum_{j'\in\mathbb{A}(i)}\exp(\langle \V{e}_i,\V{e}_{j'} \rangle_\tau)}{n}\right)+\log n
		\end{split}
	\end{equation}
	, and by continuous mapping theorem and the fact that $\exp(\langle e_i,e_j \rangle_\tau)$ is bounded, we have
	\begin{equation}
		\begin{split}
			&\lim_{n\rightarrow\infty}\log\left(\frac{\sum_{j'\in\mathbb{A}(i)}\exp(\langle \V{e}_i,\V{e}_{j'} \rangle_\tau)}{n}\right) \\
			& =\log\mathbb{E}_{j\sim p_{data}} \exp(\langle \V{e}_i,\V{e}_j \rangle_\tau)
		\end{split}
	\end{equation}
	
	Combine all the results that we have so far, we get
	\begin{equation}
		\begin{split}
			&	\lim_{n,n_k\rightarrow\infty}\left( L_i-\log n\right)\\
			&=-\mathbb{E}_{a\sim p_{pos}}\langle \V{e}_i,\V{e}_j \rangle_\tau+ \left[\log\mathbb{E}_{j\sim p_{data}} \exp(\langle \V{e}_i,\V{e}_j \rangle_\tau)\right] \\
			& +\lim_{n_k\rightarrow\infty}o(1)-\log n\nonumber\\
			&=-\mathbb{E}_{a\sim p_{pos}}\langle \V{e}_i,\V{e}_j \rangle_\tau + \log\mathbb{E}_{j\sim p_{data}} \exp(\langle \V{e}_i,\V{e}_j \rangle \tau). \\
		\end{split}
	\end{equation}
	
	The proof of Statement (1) is completed.

	\subsection{Proof of Statement (3)}
	To show the rate of convergence, it is enough to bound the following two quantities for arbitrarily small $\epsilon$:
	
	\begin{align}
		(i)&=\mathcal{P}(\large|\sum_{j\in\mathbb{P}(i)}\frac{1}{n_k}\langle \V{e}_i,\V{e}_j\rangle_\tau-\mathbb{E}_{j\sim p_{pos}}\langle \V{e}_i,\V{e}_j \rangle_\tau \large|\ge \epsilon)\nonumber\\
		(ii)& =\mathcal{P}(\large|\log(\frac{\sum_{j\in\mathbb{A}(i)}\exp(\langle \V{e}_i,\V{e}_j \rangle_\tau)}{n}) \nonumber\\
		&- \log\mathbb{E}_{j\sim p_{data}} \exp(\langle \V{e}_i,\V{e}_j \rangle_\tau) \large|\ge \epsilon)
	\end{align}
	
	By general Hoeffding's inequality and the fact that $|\langle e_i,e_a \rangle_\tau|\le \frac{1}{\tau}$, we immediately have
	\begin{align}
		(i)&\le 2\exp(-\frac{1}{2}\tau^2 n_k \epsilon^2)\nonumber\\
		&=2\exp(-\frac{1}{2}\tau^2(p_kn+o(1))\epsilon^2)\nonumber\\
		&=O(\exp(-\frac{1}{2}\tau^2 p_kn))\nonumber\\
		&=O(\exp(-Cn))
	\end{align}
	where $C$ is a universal constant.
	
	For $(ii)$, denote 
	\begin{align*}
		Z_{j}&=\exp(\langle \V{e}_i,\V{e}_j \rangle_\tau)\\
		\mu_Z&=\mathbb{E}_{j\sim p_{data}} \exp(\langle \V{e}_i,\V{e}_j \rangle_\tau)
	\end{align*}
	
	Then by Taylor's expansion of $\log$ at $\mu_Z$,
	\begin{equation}
		\begin{split}
			&	\log \left( \frac{\sum_{j \in\mathbb{A}(i)}Z_j}{n}\right) \\ & =\log\mu_Z+\frac{1}{\mu_Z}\left(\frac{\sum_{j\in\mathbb{A}(i)}Z_j}{n} -\mu_Z\right) \\
			& +o\left(\frac{\sum_{j\in\mathbb{A}(i)}Z_j}{n}-\mu_Z\right)
		\end{split}
	\end{equation}
	The first constant term is subtracted from the probability inequality. The third term can be removed to the right hand side of the inequality so that it becomes: $\epsilon-o\left(\frac{\sum_{j\in\mathbb{A}(i)}Z_j}{n}-\mu_Z\right)$. Next, we apply Hoeffding ot the second term. Since Hoeffding is a finite sample result, we let the sample size $n$ to be sufficiently large so that the third term is negligible w.r.t $\epsilon$. 
	
	Hence, similar to $(i)$, for $n$ large enough,
	\begin{align}
		(ii)&\le 2\exp(-\frac{1}{2}\tau n\mu_Z^2\epsilon^2)\nonumber\\
		&=O(\exp(-Cn))
	\end{align}
	where $C$ is a universal constant provided $\mu_Z$ is lower bounded. Thus, statement (3) is proved.

	\section{Backward Propagation Analysis in Detail}
	\label{apx: backward analyze}
	In this section, we prove that the teacher classifier and cold temperature are helpful for enlarging the effects of hard positive in backward propagation as well as making the model focus no visual similar images. First, the sum of weights in Eq.(5) in the main paper can be rewritten as: 
	\begin{align} 
		\label{equ: weightsum}
		\sum_{j{\in}\mathbb{A}(i)}w_{ij}
		&=\sum_{j \in\mathbb{P}(i)}w_{ij}+\sum_{j{\in}\mathbb{N}(i)}w_{ij}\nonumber
		\\&=\sum_{j{\in}\mathbb{P}(i)}kw+\sum_{j{\in}\mathbb{N}(i)}w
		\\&=(|\mathbb{P}(i)|k+|\mathbb{N}(i)|)w\nonumber
		\\&=Ww \nonumber
	\end{align}
	
	Then the gradient derivation of $L_{SACL}^i$ can be written as:

	\begin{flalign} 
		\label{equ: sim_derivation}
		&\frac{{\partial}L_{SACL}^i}{{\partial}\V{e}_i}&
	\end{flalign}
	\begin{flalign*}
		&=\frac{-1}{Ww}\sum_{j{\in}\mathbb{A}(i)}{w_{ij}}\{\frac{\V{e}_j}{\tau}-\frac{1}{\tau}\frac{\sum_{j'{\in}\mathbb{A}(i)}\V{e}_{j'}{\cdot}\exp(\langle \V{e}_i,\V{e}_{j'}\rangle_{\tau})}{\sum_{j'{\in}\mathbb{A}(i)}\exp(\langle
			\V{e}_i,\V{e}_{j'}\rangle_{\tau})}\}     \nonumber&
	\end{flalign*}
	\begin{flalign*}
		&=\frac{-1}{{\tau}Ww}\sum_{j{\in}\mathbb{A}(i)}{w_{ij}}\{\V{e}_j-\frac{\sum_{j'{\in}\mathbb{A}(i)}\V{e}_{j'}{\cdot}\exp(\langle \V{e}_i,\V{e}_{j'}\rangle_{\tau})}{\sum_{j'{\in}\mathbb{A}(i)}\exp(\langle \V{e}_i,\V{e}_{j'}\rangle_{\tau})}\}       \nonumber&
	\end{flalign*}
	\begin{flalign*}
		&=\frac{-1}{{\tau}Ww}\sum_{j{\in}\mathbb{A}(i)}w_{ij}\{\V{e}_j-\sum_{j'{\in}\mathbb{A}(i)}\V{e}_{j'}{\cdot}F_{ij'}\}    \nonumber&
	\end{flalign*}
	\begin{flalign*}
		&=\frac{-1}{{\tau}Ww}[\sum_{j{\in}\mathbb{P}(i)}w_{ij}\{\V{e}_j-\sum_{j'{\in}\mathbb{A}(i)}\V{e}_{j'}{\cdot}F_{ij'}\}\nonumber&
	\end{flalign*}
	\begin{flalign*}
		&+\sum_{j{\in}\mathbb{N}(i)}w_{ij}\{\V{e}_j-\sum_{j'{\in}\mathbb{A}(i)}\V{e}_{j'}{\cdot}F_{ij'}\}]     \nonumber&
	\end{flalign*}
	\begin{flalign*}
		&=\frac{-1}{{\tau}Ww}[\sum_{j{\in}\mathbb{P}(i)}kw\{\V{e}_j-\sum_{j'{\in}\mathbb{A}(i)}\V{e}_{j'}{\cdot}F_{ij'}\}\nonumber&
	\end{flalign*}
	\begin{flalign*}
		&+\sum_{j{\in}\mathbb{N}(i)}w\{\V{e}_j-\sum_{j'{\in}\mathbb{A}(i)}\V{e}_{j'}{\cdot}F_{ij'}\}]    \nonumber&
	\end{flalign*}
	\begin{flalign*}
		&=\frac{-1}{{\tau}W}[k\sum_{j{\in}\mathbb{P}(i)}\V{e}_j+\sum_{j{\in}\mathbb{N}(i)}\V{e}_j&
	\end{flalign*}
	\begin{flalign*}
		&-(|\mathbb{P}(i)|k+|\mathbb{N}(i)|)(\sum_{j'{\in}\mathbb{P}(i)}\V{e}_{j'}F_{ij'} +\sum_{j'{\in}\mathbb{N}(i)}\V{e}_{j'}F_{ij'})]\nonumber&
	\end{flalign*}
	\begin{flalign*}
		&=\frac{-1}{{\tau}W}[k\sum_{j{\in}\mathbb{P}(i)}\V{e}_j+\sum_{j{\in}\mathbb{N}(i)}\V{e}_j\nonumber-W(\sum_{j{\in}\mathbb{P}(i)}\V{e}_{j}F_{ij}+\sum_{j{\in}\mathbb{N}(i)}\V{e}_{j}F_{ij})]       \nonumber&
	\end{flalign*}
	\begin{flalign*}
		&=\frac{1}{\tau}[\sum_{j{\in}\mathbb{P}(i)}\V{e}_j(F_{ij}-\frac{k}{W})+\sum_{j{\in}\mathbb{N}(i)}\V{e}_j(F_{ij}-\frac{1}{W})]\nonumber&
	\end{flalign*}
	
	where
	\begin{equation}
		F_{ij}=\frac{\exp(\langle \V{e}_i,\V{e}_j\rangle_{\tau})}{\sum_{j'{\in}\mathbb{A}(i)}\exp(\langle \V{e}_i,\V{e}_{j'}\rangle_{\tau})}, \V{e}_i=\V{f}_i/||\V{f}_i||,\V{f}_i=f_{\phi}(\tilde{\V{x}}_i) \nonumber \nonumber
	\end{equation}
	
	\begin{align}
		\label{equ: featderiv}
		\frac{{\partial}\V{e}_i}{{\partial}\V{f}_i}=\frac{1}{||\V{f}_i||}(I-\V{e}_i\V{e}_i^T)
	\end{align}
	According to Eq.\ref{equ: sim_derivation} and \ref{equ: featderiv}, we can rewrite the formula as follow:
	\begin{align}
		\label{equ: derivationfi}
		&\frac{{\partial}L_{SACL}^i}{{\partial}\V{f}_i}
		\\=&\frac{{\partial}L_{SACL}^i}{{\partial}\V{e}_i}\frac{{\partial}\V{e}_i}{{\partial}\V{f}_i}     \nonumber\\
		=&\frac{1}{{\tau}||\V{f}_i||}(I-\V{e}_i\V{e}_i^T)[\sum_{j{\in}\mathbb{P}(i)}\V{e}_j(F_{ij}-\frac{k}{W})\nonumber+\sum_{j{\in}\mathbb{N}(i)}\V{e}_j(F_{ij}-\frac{1}{W})]      \nonumber\\
		=&\frac{1}{{\tau}||\V{f}_i||}[\sum_{j{\in}\mathbb{P}(i)}(\V{e}_j-\langle \V{e}_i,\V{e}_{j}\rangle_{1}\V{e}_i)(F_{ij}-\frac{k}{W})\nonumber\\
		&+\sum_{j{\in}\mathbb{N}(i)}(\V{e}_j-\langle \V{e}_i,\V{e}_{j}\rangle_{1}\V{e}_i)(F_{ij}-\frac{1}{W})]\nonumber
	\end{align}
	
	We now show that SACL can utilize hard positive to promote model learning from easy positive and hard negative.
	
	For an easy positive $\V{e}_j$, 
	\begin{equation}
		\langle \V{e}_i,\V{e}_{j}\rangle_{1}{\approx}1, ||\V{e}_j-\langle \V{e}_i,\V{e}_{j}\rangle_{1}\V{e}_i||=\sqrt{1-\langle \V{e}_i,\V{e}_{j}\rangle_{1}^2}\approx0. 
	\end{equation}
	
	For a hard positive $\V{e}_j$, 
	\begin{equation}
		\langle \V{e}_i,\V{e}_{j}\rangle_{1}{\approx}0, ||\V{e}_j-\langle \V{e}_i,\V{e}_{j}\rangle_{1}\V{e}_i||=\sqrt{1-\langle \V{e}_i,\V{e}_{j}\rangle_{1}^2}\approx1.
	\end{equation}
	
	Therefore, for the $\mathbb{P}(i)$ in Eq.\ref{equ: derivationfi}:
	\begin{align}
		\label{equ: pos}
		&||\V{e}_j-\langle \V{e}_i,\V{e}_{j}\rangle_{1}\V{e}_i||{\;}|F_{ij}-\frac{k}{W}|    
		\\
		&{\approx}|F_{ij}-\frac{k}{W}|     \nonumber\\
		&=|\frac{1}{\sum_{j{\in}\mathbb{A}(i)}\exp(\langle \V{e}_i,\V{e}_{j}\rangle_{\tau})}-\frac{k}{W}|    \nonumber\\
		&=|\frac{1}{\sum_{j{\in}\mathbb{P}(i)}\exp(\langle \V{e}_i,\V{e}_{j}\rangle_{\tau})+\sum_{j{\in}\mathbb{N}(i)}\exp(\langle \V{e}_i,\V{e}_{j}\rangle_{\tau})}-\frac{k}{W}|    \nonumber\\
		&{\propto}{\;}k\sum_{j{\in}\mathbb{P}(i)}\exp(\langle \V{e}_i,\V{e}_{j}\rangle_{\tau})+k\sum_{j{\in}\mathbb{N}(i)}\exp(\langle \V{e}_i,\V{e}_{j}\rangle_{\tau})-W\nonumber
	\end{align}
	The easy positive part is disappeared in gradient and  only the hard positive part preserved. It means that the contrast between an anchor image and hard positives plays an important role in model updating. Therefore, the model can learn much more from backward propagation when the anchor compares with a hard positive sample. We assume that for the easy positive and the hard negative, $\langle \V{e}_i,\V{e}_j\rangle>0$. Because both are images with high visual similarity with the anchor. Therefore, the value of $\exp(\langle \V{e}_i,\V{e}_j\rangle_{\tau})$ in Eq.\ref{equ: pos} can be large enough with the amplifying of cold temperature $\tau$. This indicates that the model majorly learning from easy positive and hard negative, which satisfying our original purpose of designing SACL to make model learn from visual similar images rather than images with the same label. Our experiments of temperatures also prove that a smaller cold temperature can help model learn better. The multiple relationship $k$ also amplifies the usage of samples in backward propagation, which means the knowledge can be transferred from classifier to feature extractor because $k$ is implied in the structure similarity information provided by classifier. 
}
\vfill

\end{document}